\documentclass[journal]{IEEEtran}

\usepackage{ifpdf}
\usepackage{graphics}
\usepackage{epsfig}
\usepackage{graphicx}

\usepackage{multirow}
\usepackage{tabularx}
\usepackage{amssymb}
\usepackage{mathtools}
\usepackage{cite}
\usepackage{hhline}
\usepackage{nopageno}
\usepackage{array}
\usepackage{float}
\usepackage{xcolor}

\usepackage[caption = false]{subfig}

\usepackage{epstopdf}
\DeclareGraphicsExtensions{.eps,.pdf}

\graphicspath{{./myfig/}}

\ifCLASSINFOpdf
\else
\fi
\usepackage{amsmath}

\begin{document}
%
\title{Anchors Based Method for Fingertips Position Estimation from a Monocular RGB Image using Deep Neural Network}
%
%
%

\author{Purnendu~Mishra,~\IEEEmembership{Student Member,~IEEE,}
        and~Kishor~Sarawadekar,~\IEEEmembership{Member,~IEEE}}
\maketitle
\thispagestyle{empty}

\begin{abstract}
\textcolor{black}{In Virtual, augmented, and mixed reality, the use of hand gestures is increasingly becoming popular to reduce the difference between the virtual and real world. The precise location of the fingertip is essential/crucial for a seamless experience. Much of the research work is based on using depth information for the estimation of the fingertips position. However, most of the work using RGB images for fingertips detection is limited to a single finger.} The detection of multiple fingertips from a single RGB image is very challenging due to various factors. In this paper, we propose a deep neural network (DNN) based methodology to estimate the fingertips position. We christened this methodology as an Anchor based Fingertips Position Estimation (ABFPE), and it is a two-step process. The fingertips location is estimated using regression by computing the difference in the location of a fingertip from the nearest anchor point. The proposed framework performs the best with limited dependence on hand detection results. In our experiments on the SCUT-Ego-Gesture dataset, we achieved the fingertips detection error of 2.3552 pixels on a video frame with a resolution of $640 \times 480$ \textcolor{black}{and about $92.98\%$ of test images have average pixel errors of five pixels.} 
\end{abstract}


%
\IEEEpeerreviewmaketitle

\section{Introduction}
%
%
%

\IEEEPARstart{H}{uman} \textcolor{black}{pose estimation is essential for many computer-vision applications such as sign language recognition, augmented reality, virtual reality, human-computer interaction, etc. In these systems, it is vital to accurate detect and track hand joints as well as fingertips. As compared to other hand joints, it is more difficult to localize the fingertips due to their higher flexibility, occlusion, etc. In this work we have focused on estimation of fingertips position in monocular RGB images.} 

The standard RGB cameras are popular, and high-resolution 2D cameras are available at relatively lower prices. The usage of a standard RGB camera in a gestural interaction system is rare.  \textcolor{black}{The methods available for estimation of fingertips position either restricted to estimation of single finger \cite{huang2015deepfinger, huang2016pointing, mukherjee2019fingertip} or lacks accuracy}. In our proposal, we used standard 2D camera to estimate the position of multiple fingertips which is a crucial step in any hand gesture-based machine interaction system. 

\begin{figure}[t]
    \centering
    \includegraphics[width=0.9\linewidth]{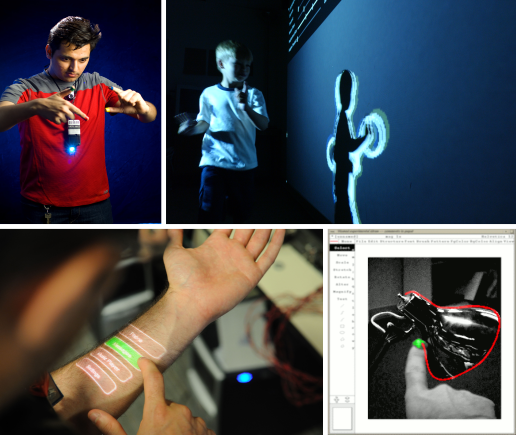}
    \caption{An illustration showing use of fingertips to interact with various object in  virtual/augmented reality environment. (Source:https://www.wikipedia.org/).}
    \label{fig:examples}
\end{figure}

Successful estimation of fingertips position can have its application in a wide variety of domains. In several augmented reality (AR) applications, hands are used to interact and manipulate the virtual content. Buchmann et al. \cite{buchmann2004fingartips} demonstrated usage hand gestures in the AR urban planning application. Thumb and index fingers were used to perform operations like pointing, grabbing, etc. Liu et al. \cite{liu2019grasp} presented an approach to grasp various objects in a robotic application by integrating the position and orientation information of thumb and wrist from a human. P. Mistry and Prattie Maes\cite{mistry2009wearable} developed a gestural interface that permits a user to interact with augmented information using marked fingertips. Billinghurst et al. \cite{billinghurst2014hands} extracted fingertips position using a depth camera and implemented numerous gesture-based interactions with virtual content in three dimensions. A finger-based writing-in-the-air system using Kinect is developed by Zhang et al. \cite{zhang2013new}. Recently, Choi et al. \cite{choi2018co} proposed a depth-based fingertip recognition method for interactive projectors. Some examples demonstrating the use of fingertips detection are shown in Fig. \ref{fig:examples}.

The challenges with fingertips detection in monocular RGB images are complex background conditions, varying illuminations, image resolutions. Moreover, the hand occupies only a small region of an image, and it becomes challenging to segment the hand area within a given RGB image \cite{modanwal2018robustwrist, modanwal2018robusthand}. Hence, it becomes difficult to estimate position of fingertips directly. \textcolor{black}{We have tried to address some of these problems in this work. We utilized abstract features learning capability of Convolutional Neural Networks (CNNs) and have followed the two-step approach for multiple fingertips detection. First step is to detect the hand area using a deep neural network (DNN) based object detection model.  Having obtained the hand region, in the second step a few fixed (known) points called \lq anchors\rq{ }are placed on this segmented image. With the help of these anchors, nearest fingertips to them are identified and then their positions are estimated.}   \textcolor{black}{The contribution of this paper are summarized as follows:
\begin{itemize}
    \item We propose an anchor-based approach for the estimation of fingertips position in a multi-gesture scenario where different combination of fingers are used to make a hand pose. A fingertip is detected by finding the nearest anchor point and then estimating the offset from that anchor location.
    \item We demonstrated the effectiveness of anchor-based method for estimation of fingertip position on single RGB image. The use of anchor locations have assisted in identifying  fingertip location of multiple fingers present in the given image and reducing the detection error.
\end{itemize}
}

The organization of the rest of the paper is as follows. A brief overview of the literature available on the estimation of fingertip position is given in Section \ref{sec:related}.  The proposed methodology is discussed in Section \ref{sec:details}.  The details of the experiment are given in Section \ref{sec:experiment}. In Section \ref{sec:results}  experimental results are discussed, and finally, Section \ref{sec:conclusion} concludes the paper.

\section{Related Works}
\label{sec:related}

In this section, we discuss research work available in the field of fingertips position estimation. Research work available in this area can be categorized as non-vision based methods and vision-based methods. The non-vision methods use hand gloves to locate fingertips position. Sensors mounted on fingers are used to track the movement of fingers. The information received from sensors is used for interactive operation in a virtual environment.  Buchmann et al. \cite{buchmann2004fingartips} used a glove as an input device to interact with the 3D graphics. The hand is rendered in the virtual urban planning workspace, and operations like grabbing, pointing, pressing, and navigation are performed. In related work, Dorfmuller-Ulhaas et al. \cite{dorfmuller2001finger} and Zhang et al. \cite{zhang2005interaction} accessed fingertips position using data gloves. The rewarding point of data gloves method is the improved accuracy of the system. However, hand gloves restrict hand movement, and it is inconvenient for users too. Due to these limitations, the researchers have focused on vision-based methods.

Vision-based fingertips detection methods involve the use of unique markers. The fingertips are either marked with paint or colored tapes. It is easy to track these markers and estimate the position of the fingertips. Nakamura et al. \cite{nakamura2008double} tracked fingertips by placing an LED light source on them. In the Sixth Sense device \cite{mistry2009wearable}, the user's fingers are taped with multiple color markers.  The system tracks a specific color to identify the finger, and the user interacts with the projected contents. However, it is very inconvenient for the user to put a marker every time. Hence, more research is needed in vision-based fingertips detection without causing any inconvenience to the user.

\begin{figure}[t]
    \centering
    \includegraphics[width=\linewidth]{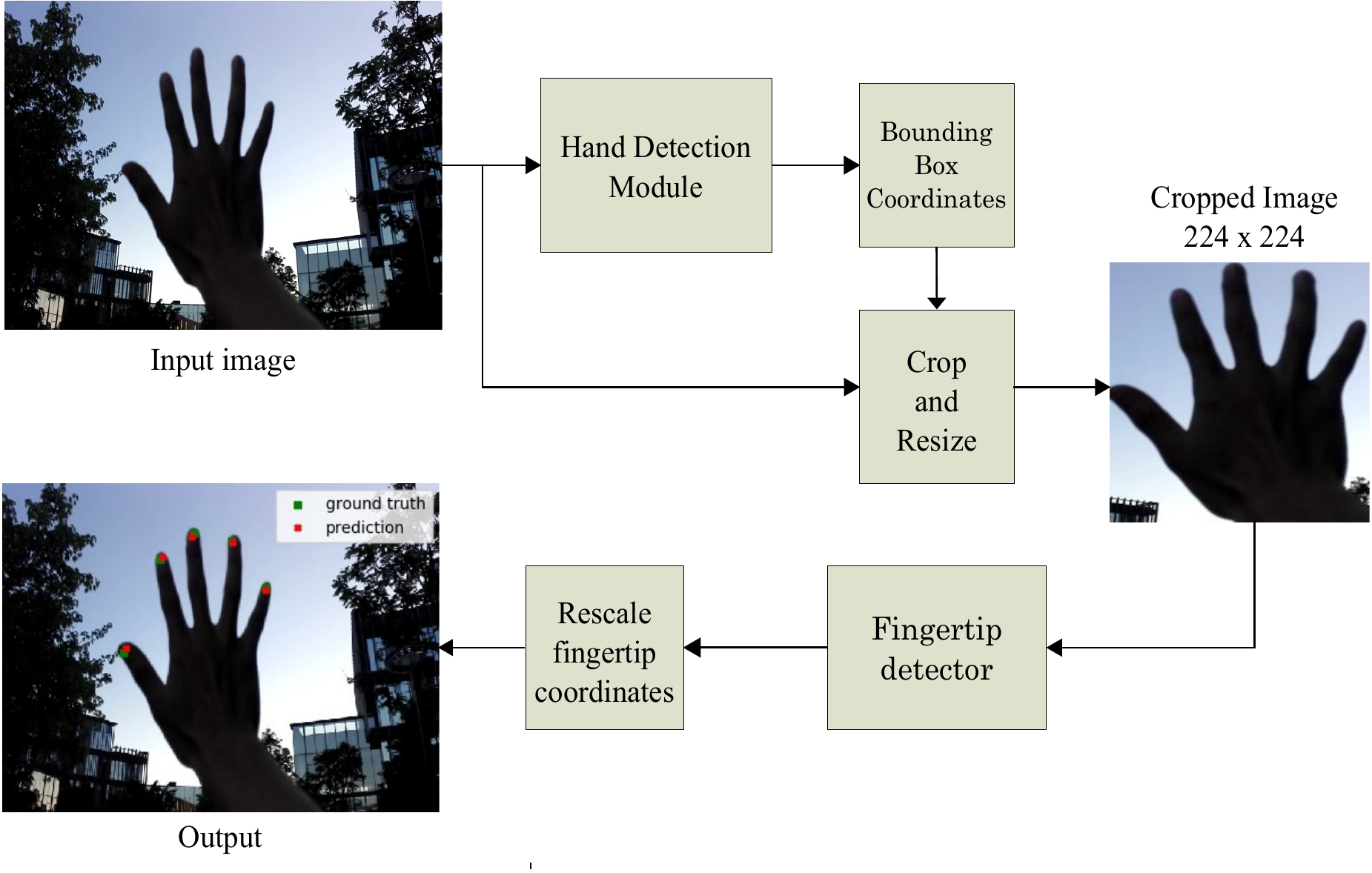}
    \caption{The proposed framework for the estimation of fingertip(s) position.}
    \label{fig:framework}
\end{figure}

The process of estimation of fingertips position in the appearance-based method, without using any special markers, in general, is divided into two steps: hand area detection and locating fingertips. Hand region localization is the first step in any hand-based application, and it affects the performance of the subsequent procedures. The commonly used method for achieving hand localization is skin color segmentation as color differences between the human skin tone and the background is distinguishable. The skin color-based methods utilize different color spaces for hand localization. YCbCr color space is used by Lee et al. \cite{lee2012robust}, and G. Wu et al. \cite{wu2016robust} for hand region detection. The color-based models face the challenge of color similarity, lighting variations, and background complexity \cite{modanwal2018robustwrist,zhang2013new}. Depth-based hand segmentation is unaffected by these issues. Several methods adopted this technique \cite{choi2018co, liang2014parsing, zhang2013new, ren2013robust, wang2015superpixel,modanwal2018robusthand} for hand-detection. However, drawbacks of this method are - noise affects the depth information. It has a small range, and it is ineffective in outdoor conditions \cite{zhang2013new, wu2017yolse, wang2020srhandnet}. Recently CNN based models are extensively used for object detection and localization. Y. Huang et al. \cite{huang2016pointing} and Mukherjee et al. \cite{mukherjee2019fingertip} used Faster-RCNN \cite{ren2015faster} to localize the hand in the RGB image in their respective work. The work of Y. Huang et al. \cite{huang2015deepfinger}, Mukherjee et al. \cite{mukherjee2019fingertip}, and W. Wu et al. \cite{wu2017yolse} rely on CNN based object detector to obtain bounding boxes around the hand region. In the proposed work, we rely on the YOLOv3 (You Only Look Once) model \cite{redmon2018yolov3} for hand area segmentation.

The next step after the hand localization is to estimate the fingertips position. The conventional methods are K curvature method, fingertip template matching, focus distance method, contour finding \cite{baldauf2011markerless, lee2011vision, kulshreshth2013poster, lee2012robust, wu2016robust}. Wu et al. \cite{wu2016robust} proposed average centroid distance (ACD) method and  Mukherjee et al. \cite{mukherjee2019fingertip} proposed the distance weighted curvature entropy method. These methods are dependent on the robustness of skin segmentation process. The method proposed by Mukherjee et al. \cite{mukherjee2019fingertip} works only for a single finger. Some researchers have started exploiting complex feature extraction capabilities of CNN for fingertips position estimation. Y. Huang et al. \cite{huang2016pointing, huang2015deepfinger}, use CNN to find fingertips location. But, their model is only capable of locating the fingertip of a single finger. A method using CNN, which relies on heat-map to locate multiple fingertips, is proposed by Wu et al. \cite{wu2017yolse}. Although this method is effective in locating multiple fingertips, it lacks accuracy, and sometimes the heat-map prediction has a few false-positives. We have proposed an improvement over the method in \cite{wu2017yolse} called as Fingertips Position Estimation with Fingers Identification (FPEFI) \cite{purnendu2019fingertips} to estimate the fingertips position. It uses regression combined with multi-label classification to locate the fingertips in an RGB image. The FPEFI method needs hand gesture information to detect the fingertips. In this work, we remove that constraint and estimate the fingertips position.

\section{The Proposed Method}
\label{sec:details}
 The framework of the proposed methodology for fingertip(s) detection is shown in Fig. \ref{fig:framework}. \textcolor{black}{The process followed in the proposed work is similar to our previous work for fingertip detection in  \cite{purnendu2019fingertips}. The step of hand detection is same in the current and  the previous work. The difference is in the methodology of fingertips detection. For hand area segmentation we use a DNN based object detection method. The hand area is extracted from the input image using the bounding box coordinates obtained from the DNN model. The image thus obtained is used as input to the fingertip(s) detection module.}

\textcolor{black}{The methodology of fingertips detection in \cite{purnendu2019fingertips}, we used direct regression to estimate the fingertip location on cropped image. In addition, we performed multi-label classification to identify the extended finger used in hand pose. We estimated fingertips position by combining these outputs. Different from method in \cite{purnendu2019fingertips}, here we propose to estimate the fingertips position with the help fixed anchor points. The process is named as Anchor based Fingertips Position Estimation (ABFPE). The location of a fingertip is calculated as an offset from the nearest anchor point. The anchor point has advantage of predicting multiple fingertip location with varying hand pose and a fingertip location is obtained with improved accuracy then method present in \cite{purnendu2019fingertips}.}

The process of hand detection and the fingertip(s) position estimation is described below.


\begin{figure}[!b]
\centering
\subfloat[Gesture two]{\includegraphics[width = 0.5\linewidth]{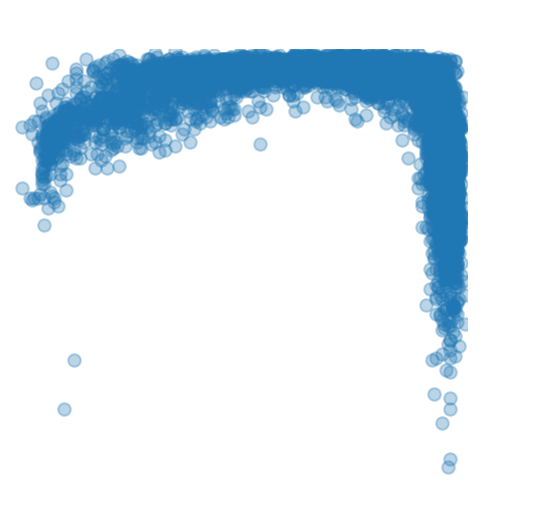}} 
\subfloat[Gesture three]{\includegraphics[width = 0.5\linewidth]{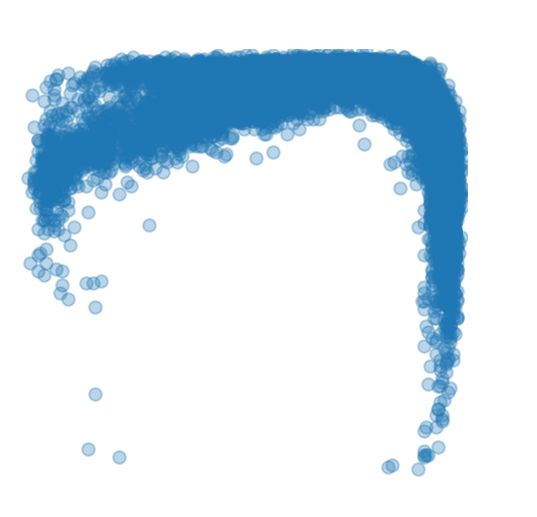}}\\
\subfloat[Gesture four]{\includegraphics[width = 0.5\linewidth]{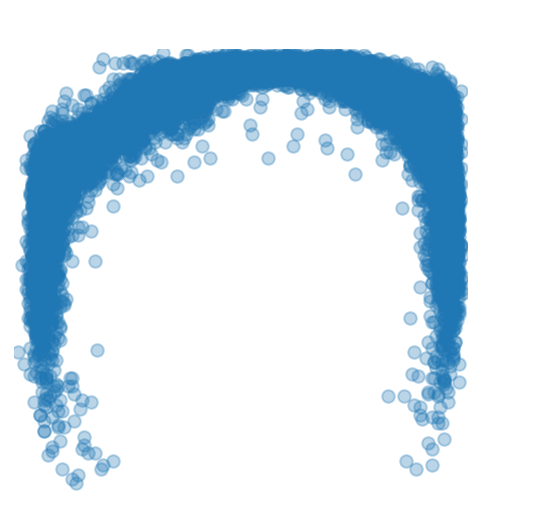}}
\subfloat[Gesture five]{\includegraphics[width = 0.5\linewidth]{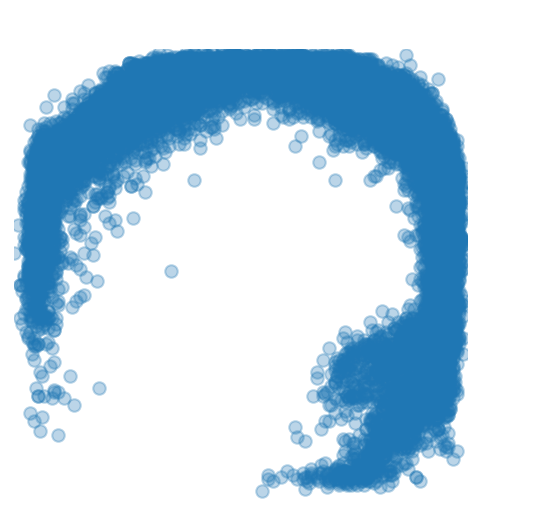}}
\caption{
The scatter plot showing the distribution of fingertips location for various gestures made using single hand. The data of fingertip(s) location is from SCUT-Ego-Gesture \cite{wu2017yolse} dataset.}
\label{fig:distribution}
\end{figure}

\begin{figure}[!t]
    \centering
    \includegraphics[width=1.0\linewidth]{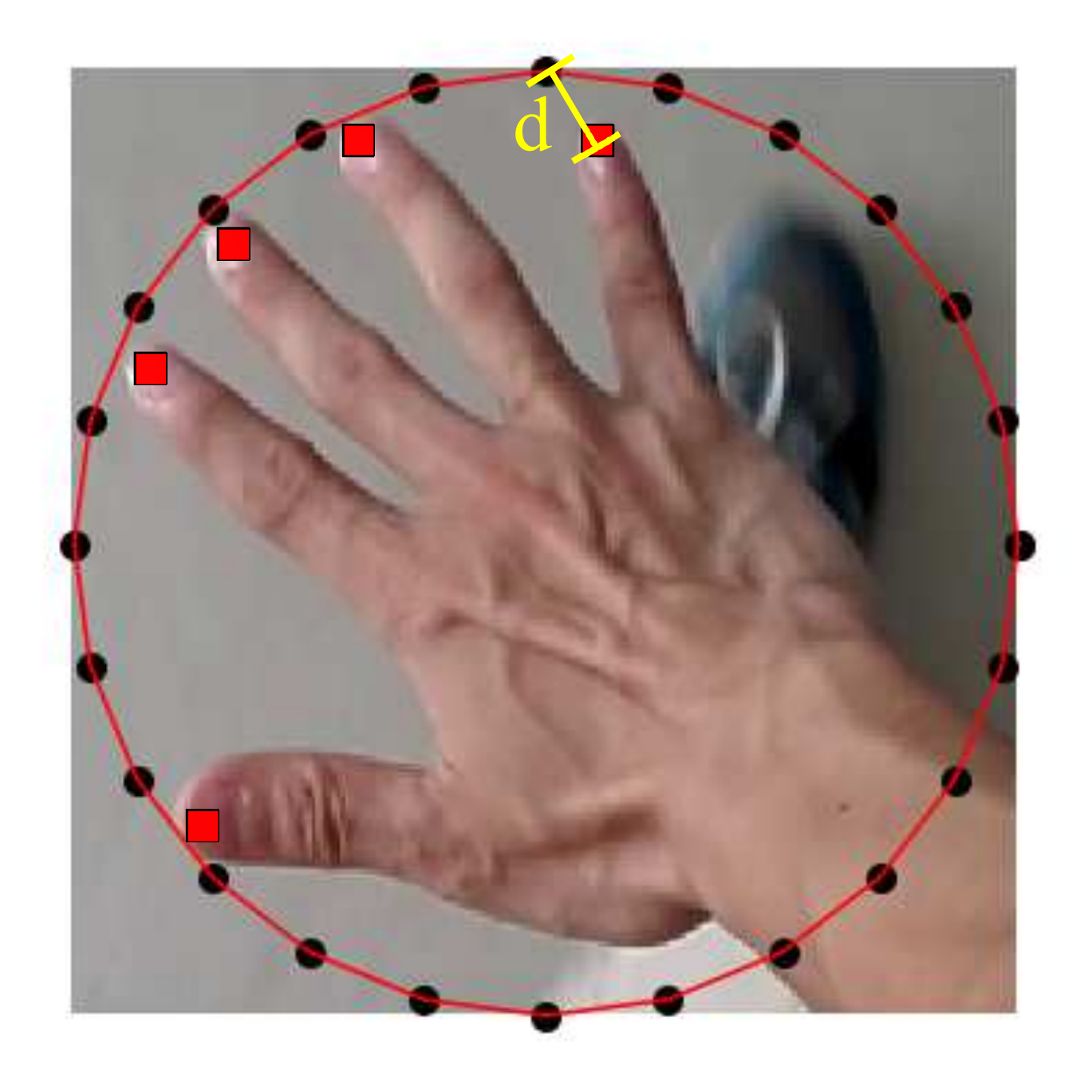}
    \caption{Instead of using regression to directly estimate the positions of fingertips (marked with red square boxes), we estimate the error \lq d\rq{ } of a fingertip from the nearest anchors point (marked with black dot).}
    \label{fig:anchors}
\end{figure}

\subsection{Hand Segmentation}
A hand occupies relatively small region in an image. Also, factors-like complex background, illumination variations, presence of human skin like objects in the scene, camera characteristics, etc. affect the hand segmentation process. Since the process of fingertip(s) position estimation, gesture recognition, hand pose estimation, etc. depends on the robustness of a hand detection process, the algorithm used to detect hand must be robust. Recently, DNN based object detection methods have shown their robustness to multiple object detection in an image. Some of the popularly used DNN architectures for object detection are Faster-RCNN \cite{ren2015faster}, YOLO \cite{redmon2017yolo9000} and SSD (Single Shot multi-box Detector) \cite{liu2016ssd}. The performance of these multiple object detection architectures is significantly better in comparison to the other methods used for the hand segmentation process. Moreover, these methods work extremely fast. In the proposed work, we re-trained YOLOv3 \cite{redmon2018yolov3} on SCUT-Ego-Gesture dataset \cite{wu2017yolse} for hand detection. In order to compare the performance of YOLOv3 for hand area detection, three different CNN architectures were trained on the aforesaid dataset. The first architecture is proposed by Y. Huang et al. \cite{huang2015deepfinger} for hand segmentation. The architecture called \lq\lq DeepFinger\rq\rq{ }uses regression technique, and it generates bounding box coordinates for the detected hand area. The other two architecture uses ResNet50 model \cite{he2016deep} for feature extraction. We replaced the classification layer in the original ResNet50 model with a linear layer comprising of four output nodes for bounding box regression. The only difference between the two ResNet50 based models is that the first model uses direct regression to generate the bounding box around the detected hand area while the second the model uses a prior (a rectangular box of dimension $200 \times 300$). The outputs of the second model are offset to the height and the width of the prior. The aforesaid second method uses a similar concept as being used in YOLO \cite{redmon2017yolo9000} and SSD \cite{liu2016ssd}. The performance comparison of various methods is detailed in Section \ref{sec:results}.

\begin{figure*}[!t]
    \centering
    \includegraphics[width=0.9\linewidth]{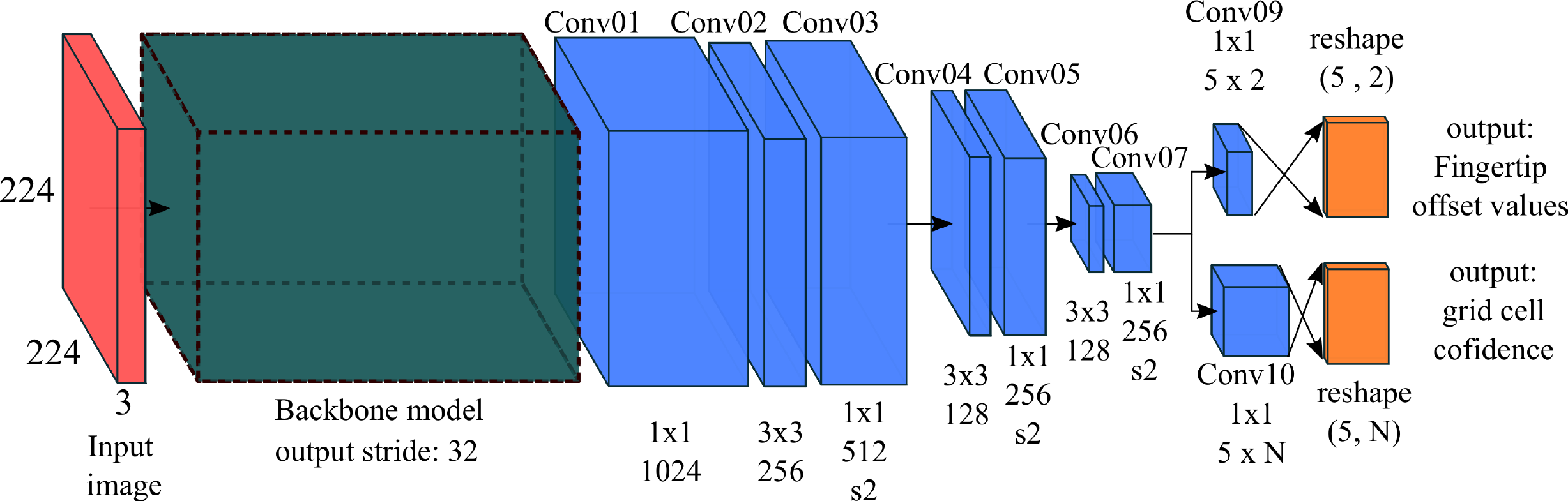}
    \caption{The architecture used for estimation of fingertips position. The backbone model can be any state-of-the-art model (e.g. VGG16 \cite{vgg16}, ResNet \cite{he2016deep}, etc.).}
    \label{fig:model}
\end{figure*}

\subsection{Fingertips Detection}
\textcolor{black}{The output of hand detection module are bounding box coordinates enclosing the hand area. These bounding box tightly enclose the hand area in the input image. And the fingertips of all the extended fingers lies near the edges of these bounding box. To verify this hypothesis, we analyzed the distribution of fingertip(s) location from the SCUT-Ego-Gesture \cite{wu2017yolse} dataset. It is observed from the scatter plot of the fingertips distribution in the bounding box region that the fingertips are clustered near the edges of the bounding box enclosing the hand.} The scatter plot of fingertips distribution for gesture two, three, four, and five are shown in Fig. \ref{fig:distribution}. Majority of fingertips lie near the edges of the bounding box, as shown in Fig. \ref{fig:distribution}. 

Based on this observation, we can search for fingertip(s) along the edges of the bounding box instead of searching it in the entire image. The proposed methodology for the fingertip(s) position estimation is explained with the help of Fig. \ref{fig:anchors}. Gesture five is posed in this figure. The hand area is cropped from the original image using the bounding box coordinates. \textcolor{black}{The cropped RGB image is then resized to $h \times w$ dimension. In the proposed work we use 224 as values for $h$ and $w$.} The ground truth fingertips are marked with red squares. Next, some points are fixed at regular intervals, along the circumference of the circle but on the edges of the image. We call these points \lq\lq anchors.\rq\rq{ }The black points in Fig. \ref{fig:anchors} represent these anchor points. The position of a fingertip is estimated by identifying the nearest anchor to the fingertip. Then distance between the anchor and the finger is estimated. For example, as shown in Fig. \ref{fig:anchors}, the Pinky finger is at distance of \textit{d} from the nearest anchor. Further, we estimate the value of \textit{d} w.r.t. the nearest anchor and obtain position of the fingertip of the Pinky finger. 

\textcolor{black}{Suppose $X_k \in \mathbb{R}^2$ where $k \in [1,5]$ represents the location of fingertip in the given RGB image. And the anchors point are represented as $Z_n \in \mathbb{R}^2$ where $n \in [1,N]$. Here $N$ is the number of anchor points. For a fingertip location $X_i$ the value of $d$ from the nearest anchor point $Z_j$ can be calculated as}

\begin{equation}
\color{black}
    d = \| Z_j - X_i\|
\end{equation}

\textcolor{black}{To identify the nearest anchor point and estimate the value for $d$ we have have used CNN. The former is treated as classification problem and for later case we use regression technique. During training of CNN, the anchor position is one-hot encoded.}

\textcolor{black}{The architecture to implement the proposed method for the estimation of fingertips position is shown in Fig. \ref{fig:model}. The network design is very simple and the details of architecture are given as follows. The network has a backbone model that extracts features from the RGB image. We experimented with three different state-of-the-art DNN models as the backbone model. These models were VGG16 \cite{vgg16}, ResNet50 \cite{he2016deep} and MobileNetV2 \cite{sandler2018mobilenetv2}. These models were initialized with pre-trained weights on the ImageNet dataset \cite{krizhevsky2012imagenet}. The backbone model has an output stride of 32. The feature map obtained from the backbone model is passed through a series of convolutional layers and its spatial resolution is reduced to $1 \times 1$. This feature map will then acts as input to two sub-networks. The first sub-networks estimate the offset value for each finger and  has a convolutional layer with 10 filters which is reshaped to an array of shape (5,2). This array stores the difference in the values of $x$ and $y$ coordinates for each finger w.r.t. the nearest anchor coordinate. The second sub-network identifies out of the $N$ anchors, which is nearest to a particular finger in the input image. We considered $N = 24$ in our experiment. This value is empirically chosen. The output of the second network is an $(5,N)$ array where each row identifies nearest anchor for respective fingertip location.}

\section{Experiments}
\label{sec:experiment}

\subsection{Dataset}
We experimented with three different publicly available datasets to validate the proposed theory. These are SCUT-Ego-Gesture dataset \cite{wu2017yolse}, SCUT-Ego-Finger dataset \cite{huang2015deepfinger, liu2015fingertip} and HGR \cite{kawulok2014self, nalepa2014fast, grzejszczak2016hand} dataset. We have selected single hand gesture images from the SCUT-Ego-Gesture dataset. The dataset has 40,636 RGB images of resolution $640 \times 480$ in the egocentric view. The dataset has normalized coordinates for the bounding box, the fingertips, and the finger joints for the outstretched fingers. We selected 32,508 images for training and 8,128 images for testing from the dataset.

The performance of the proposed fingertips position estimation algorithm has been tested on the SCUT-Ego-Finger dataset and HGR dataset. The SCUT-Ego-Finger contains hand images in an egocentric view with only the index finger of the right hand in the extended position. The proposed model is fine-tuned on 17,128 images, and 68,513 images are used to test the performance of the model. Similarly, the proposed model is fine-tuned and tested on 927 and 309 images for the HGR dataset, respectively. The HGR dataset provide annotation for 27 points of a hand. During our experiments, we selected only those annotations in which extended fingers are used to pose a gesture. All aforesaid datasets contain images with illumination variation, complex background, skin-like objects in the background, making fingertips position estimation process very challenging. 

\subsection{Data augmentation}
The proposed method for estimation of fingertips position does not take into account the hand pose, and it is robust to hand rotation. Therefore, the network is trained by augmenting the dataset by flipping hand randomly both vertically and horizontally. Random rotation of image is also performed.

\subsection{Loss Function}
The network performs two operations. First, it finds the nearest anchor point corresponding to a fingertip, and the second operation is to estimate the error between the position of the anchor point and the fingertip. Hence, the optimization of the network is achieved by utilizing two different loss functions. The loss function used during training of the proposed network is defined as
\begin{equation}
    loss = L_{classification} + L_{regression}.
\end{equation}
Here, $L_{classification}$ is cross-entropy and $L_{regression}$ is Huber loss \cite{huber1992robust}. The $L_{classification}$ loss is used for estimating the anchor nearest to a fingertip. During training, the ground truth anchor positions were one-hot encoded. The $L_{regression}$ loss is used to minimize the error between estimated position of a fingertip with respect to the nearest estimated anchor point.

\subsection{Optimization}
The complete network is trained end-to-end. The network's loss is optimized using Nesterov's accelerated stochastic gradient descent algorithm. The momentum value is set to 0.9. The network is trained using a polynomial learning rate policy with a warm restart \cite{purnendu2019polynomial}. It converges faster in comparison to \lq\lq step\rq\rq{ } learning rate annealing policy, and the network yields better performance for the same number of iterations. The \lq\lq poly\rq\rq{ }learning rate policy is defined as
\begin{equation}
    lr = lr_0 * \left( 1 - \frac{i}{M} \right)^{power}
\end{equation}
where $lr_0$ is the initial learning rate, $i$ is the current iteration value and $M$ is the maximum number of iterations during training. The value of power is set to 0.9. The warm restart takes place at a fraction of $25\%$ of the total number of epochs. The value of the initial learning rate and batch size was $10^{-2}$ and 64, respectively. The value of the learning rate at the restart was $6.5 \times 10^{-3}$. The network was trained for 100 epochs. The model was implemented using Keras \cite{chollet2015keras} with Tensorflow backend. All the experiments were performed on \textit{Nvidia GeForce GTX  1080 GPU} with 8 GB of VRAM.


\begin{figure}[!hbp]
    \centering
    \includegraphics[width=0.9\linewidth]{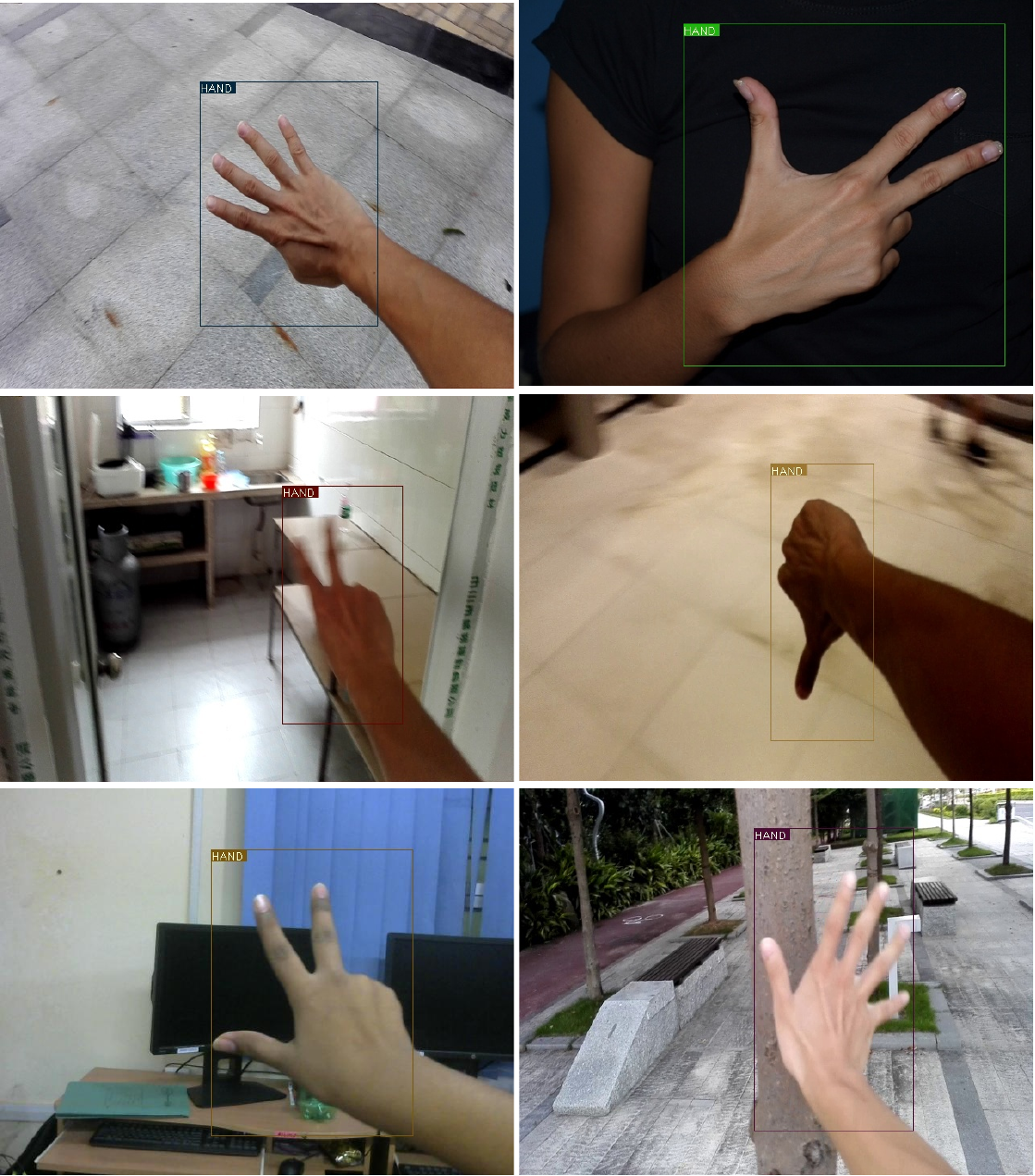}
    \caption{Some instances of hand region detection using the YOLOv3 \cite{redmon2018yolov3} model.}
    \label{fig:hand_detect}
\end{figure}

\begin{table}[!b]
\centering
\caption{Performance on hand detection on SCUT-Ego-Gesture \cite{wu2017yolse} dataset. }
\label{tab:hand-detect}
\resizebox{\linewidth}{!}{%
\begin{tabular}{|c|l|c|c|}
\hline
                          &                   & Architectures & IoU        \\ \hline
-                         &                   & DeepFinger \cite{huang2015deepfinger}   & 0.7693          \\ \hline
\multirow{2}{*}{modified} & direct regression & ResNet50      & 0.8061          \\ \cline{2-4} 
                          & using prior      & ResNet50      & 0.8406          \\ \hline
retrained                 &                   & YOLOv3   \cite{redmon2018yolov3}    & \textbf{0.9046} \\ \hline
\end{tabular}%
}
\end{table}

\begin{table*}[!t]
\centering
\tiny
\caption{Comparative results of the proposed method for various performance metrics on SCUT-Ego-Gesture \cite{wu2017yolse} dataset with three different state-of-the-art DNN architectures as backbone.}
\label{tab:backbones}
\resizebox{\linewidth}{!}{%
\begin{tabular}{|c|c|c|c|c|c|c|c|}
\hline
\begin{tabular}[c]{@{}c@{}}Backbone\\ model\end{tabular} & Image Size & \begin{tabular}[c]{@{}c@{}}Avg. Pixel \\ error (px)\end{tabular} & \begin{tabular}[c]{@{}c@{}}Threshold\\ (px)\end{tabular} & Precision & Recall & $F_{1}score$ & \begin{tabular}[c]{@{}c@{}}Time\\ (ms)\end{tabular} \\ \hhline{========}
\multirow{2}{*}{VGG16 \cite{vgg16}} & \multirow{6}{*}{$640 \times 480$} & \multirow{2}{*}{9.2573} & 10 & 0.9385 & 0.8705 & 0.8947 & 15.27 \\ \cline{4-8} 
 &  &  & 15 & 0.9738 & 0.9396 & 0.9518 & 15.26 \\ \cline{1-1} \cline{3-8} \hhline{=~======}
\multirow{2}{*}{ResNet50 \cite{he2016deep}} &  & \multirow{2}{*}{\textbf{2.3552}} & 10 & 0.9921 & 0.9824 & 0.9859 & 16.38 \\ \cline{4-8} 
 &  &  & 15 & 0.9976 & 0.9920 & 0.9940 & 16.44 \\ \cline{1-1} \cline{3-8} \hhline{=~======}
\multirow{2}{*}{MobileNetv2 \cite{sandler2018mobilenetv2}} &  & \multirow{2}{*}{3.4483} & 10 & 0.9875 & 0.9716 & 0.9772 & 10.77 \\ \cline{4-8} 
 &  &  & 15 & 0.9954 & 0.9892 & 0.9914 & 10.83 \\ \hline
\end{tabular}%
}
\end{table*}

\begin{table*}[!t]
\centering
\caption{Comparative results for various performance metrics on SCUT-Ego-Gesture \cite{wu2017yolse} dataset.}
\label{tab:egogesture}
\resizebox{\textwidth}{!}{%
\begin{tabular}{|c|c|c|c|c|c|c|c|c|c|}
\hline
Algorithm & Backbone Model & Probability & \begin{tabular}[c]{@{}c@{}}Actual \\ fingertips\end{tabular} & \begin{tabular}[c]{@{}c@{}}Detected \\ fingertips\end{tabular} & \begin{tabular}[c]{@{}c@{}}Avg. Pixel \\ error (px)\end{tabular} & Precision & Recall & $F_1score$ & \begin{tabular}[c]{@{}c@{}}Time \\ (ms)\end{tabular} \\ \hhline{==========}
\multirow{2}{*}{SpatialNet \cite{pfister2015spatialnet}} & - & 0.2 & \multirow{9}{*}{18554} & 18351 & 6.9388 & 0.9509 & 0.9452 & 0.9422 & 40.894 \\ \cline{2-3} \cline{5-10} 
 & - & 0.5 &  & 16748 & 9.8587 & 0.9343 & 0.8816 & 0.8991 & 40.895 \\ \cline{1-3} \cline{5-10} \hhline{===~======}
\multirow{2}{*}{YOLSE \cite{wu2017yolse}} & - & 0.2 &  & 19229 & 8.6793 & 0.9617 & 0.9852 & 0.9702 & 9.018 \\ \cline{2-3} \cline{5-10} 
 & - & 0.5 &  & 18386 & 10.9832 & 0.9394 & 0.9746 & 0.9521 & 9.067 \\ \cline{1-3} \cline{5-10} \hhline{===~======}
\multirow{2}{*}{FPEFI \cite{purnendu2019fingertips}} & MobileNetV2 & - &  & 18541 & 6.1527 & 0.9711 & 0.9914 & 0.9785 & 9.072 \\ \cline{2-3} \cline{5-10} 
 & ResNet50 & - &  & 18544 & 4.8224 & \textbf{0.9977} & \textbf{0.9925} & \textbf{0.9944} & 17.453 \\ \cline{1-3} \cline{5-10}  \hhline{===~======}
\multirow{3}{*}{\begin{tabular}[c]{@{}c@{}}Proposed \\ (ABFPE)\end{tabular}} & VGG16 &  &  & 18426 & 9.2573 & 0.9738 & 0.9396 & 0.9518 & 15.265 \\ \cline{2-3} \cline{5-10} 
 & MobileNetV2 & - &  & 18548 & 3.4483 & 0.9954 & 0.9892 & 0.9914 & 10.829 \\ \cline{2-3} \cline{5-10} 
 & ResNet50 & - &  & 18544 & \textbf{2.3552} & 0.9976 & 0.9920 & 0.9940 & 16.443 \\ \hline
\end{tabular}%
}
\end{table*}

\begin{table*}[!htbp]
\centering
\caption{Comparative results for various performance metrics on HGR \cite{grzejszczak2016hand} dataset.}
\label{tab:hgr}
\resizebox{\textwidth}{!}{%
\begin{tabular}{|c|c|c|c|c|c|c|c|c|c|}
\hline
Algorithm & Backbone Model & Probability & \begin{tabular}[c]{@{}c@{}}Actual \\ fingertips\end{tabular} & \begin{tabular}[c]{@{}c@{}}Detected \\ fingertips\end{tabular} & \begin{tabular}[c]{@{}c@{}}Avg. Pixel \\ error (px)\end{tabular} & Precision & Recall & $F_1score$ & \begin{tabular}[c]{@{}c@{}}Time \\ (ms)\end{tabular} \\ \hhline{==========}
SpatialNet \cite{pfister2015spatialnet} & - & 0.2 & \multirow{7}{*}{927} & 943 & 18.8543 & 0.9164 & 0.9290 & 0.9108 & 38.235 \\ \cline{1-3} \cline{5-10} \hhline{===~======}
YOLSE \cite{wu2017yolse} & - & 0.2 &  & 992 & 28.1352 & 0.8411 & 0.8613 & 0.8267 & 9.047 \\ \cline{1-3} \cline{5-10} \hhline{===~======}
\multirow{2}{*}{FPEFI \cite{purnendu2019fingertips}} & MobileNetV2 & - &  & 874 & 26.6788 & 0.8614 & 0.6582 & 0.7195 & \textbf{8.756} \\ \cline{2-3} \cline{5-10} 
 & ResNet50 & - &  & 891 & 22.4235 & 0.9352 & 0.7715 & 0.8240 & 13.115 \\ \cline{1-3} \cline{5-10} \hhline{===~======}
\multirow{3}{*}{\begin{tabular}[c]{@{}c@{}}Proposed \\ (ABFPE)\end{tabular}} & VGG16 & - &  & 915 & 14.0069 & 0.9595 & 0.8734 & 0.9026 & 14.076 \\ \cline{2-3} \cline{5-10} 
 & MobileNetV2 & - &  & 930 & 12.9178 & 0.9636 & 0.9055 & 0.9220 & 10.776 \\  \cline{2-3} \cline{5-10} 
  & ResNet50 & - &  & \textbf{925} & \textbf{8.8267} & \textbf{0.9836} & \textbf{0.9104} & \textbf{0.9342} & 16.245 \\ \hline
\end{tabular}%
}
\end{table*}

\begin{figure*}[!thb]
    \centering
    \subfloat[SpatialNet \cite{pfister2015spatialnet}]{\includegraphics[width=\textwidth]{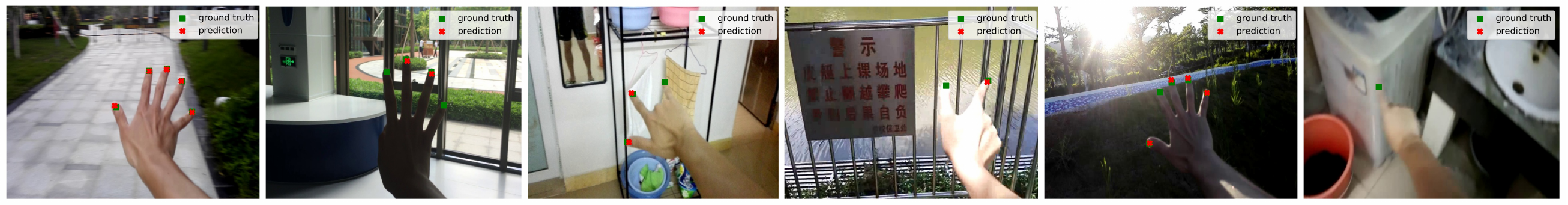}}\\
    \subfloat[YOLSE \cite{wu2017yolse}]{\includegraphics[width=\textwidth]{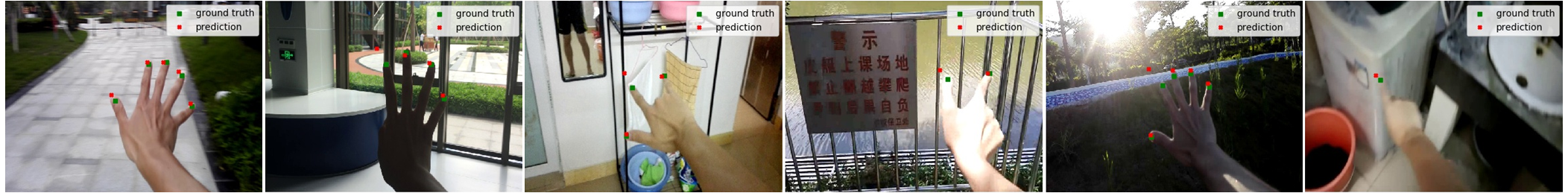}}\\
    \subfloat[FPEFI \cite{purnendu2019fingertips}]{\includegraphics[width=\textwidth]{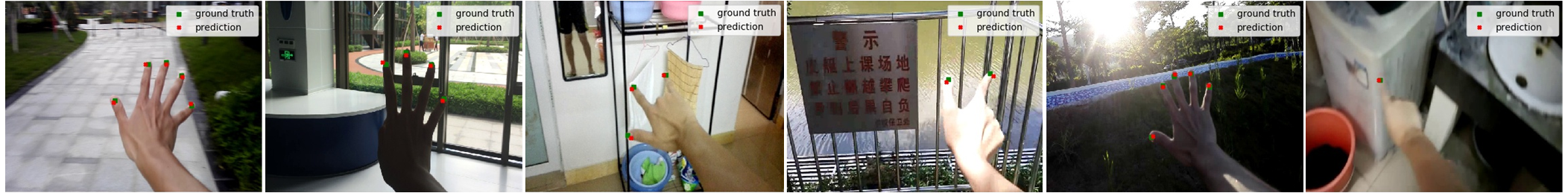}}\\
    \subfloat[The proposed method (ABFPE) ]{\includegraphics[width=\textwidth]{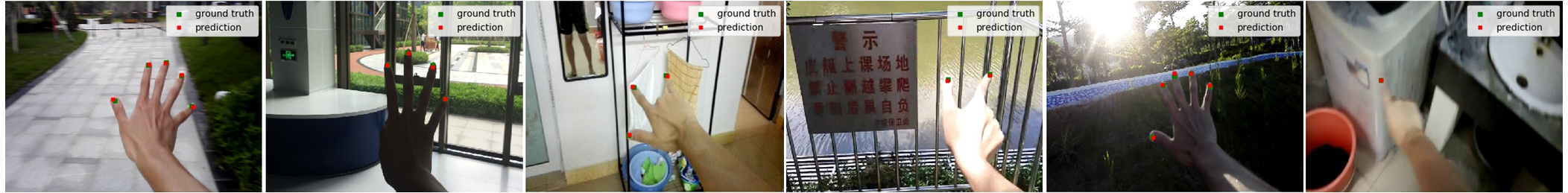}}
    \caption{Some samples results of fingertips detection results on SCUT-Ego-Gesture \cite{wu2017yolse} dataset. }
    \label{fig:egogesture}
\end{figure*}

\begin{figure*}[!tbh]
    \centering
    \subfloat[SpatialNet \cite{pfister2015spatialnet}]{\includegraphics[width=\textwidth]{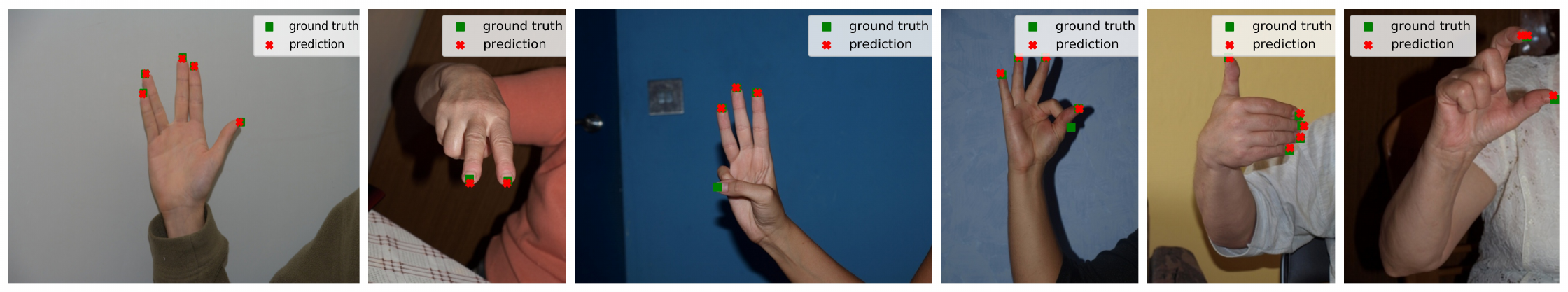}}\\
    \subfloat[YOLSE \cite{wu2017yolse}]{\includegraphics[width=\textwidth]{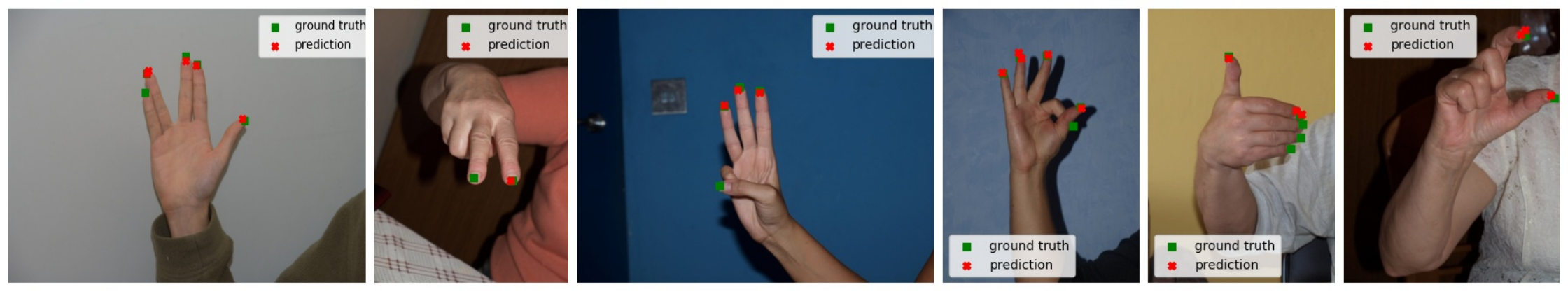}}\\
    \subfloat[FPEFI \cite{purnendu2019fingertips}]{\includegraphics[width=\textwidth]{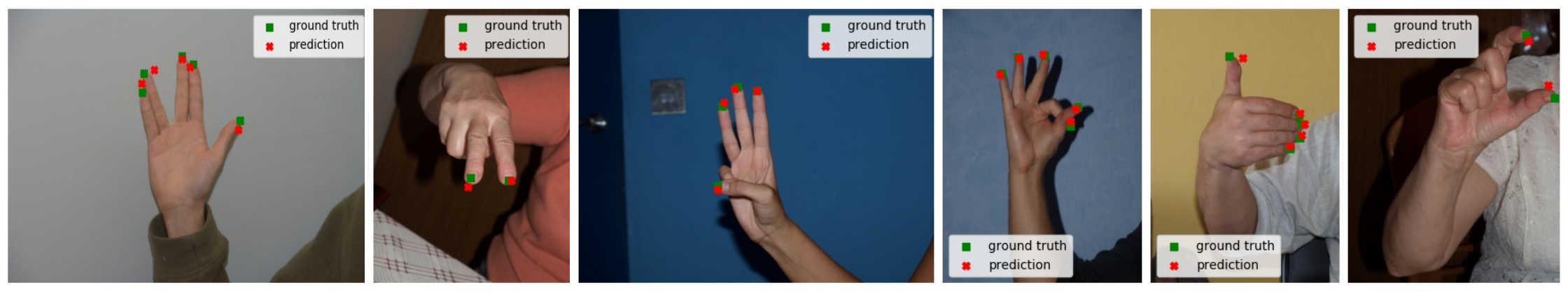}}\\
    \subfloat[The proposed method (ABFPE)]{\includegraphics[width=\textwidth]{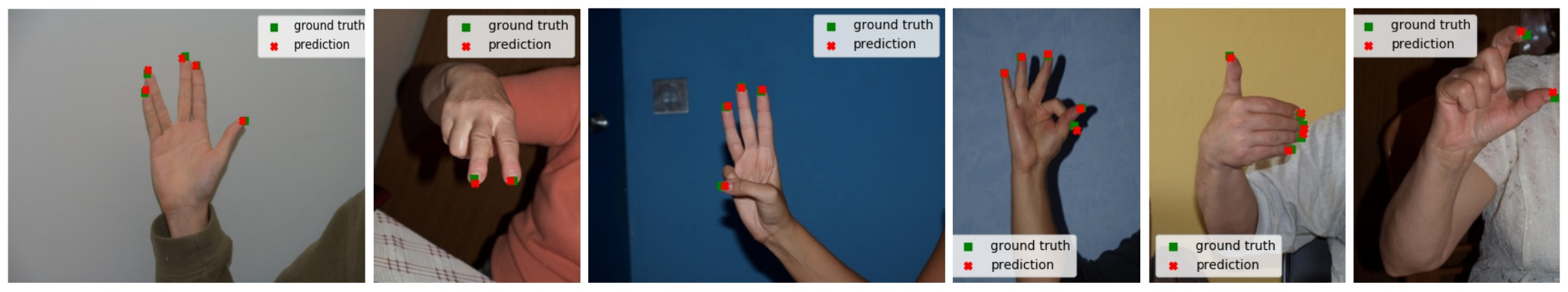}}
    \caption{Some samples results of fingertips detection results on HGR \cite{grzejszczak2016hand} dataset.}
    \label{fig:hgr}
\end{figure*}

\section{Results and Discussion}
\label{sec:results}
\subsection{Hand detection}
In general, the process of the estimation of fingertips position or the estimation of hand pose is carried out by first segmenting the hand region. Then the next step is locating fingertips. Similar approach is followed in the proposed work to estimate fingertips position from a single RGB image. The framework of the proposed methodology is shown in Fig. \ref{fig:framework}. The first step of hand region segmentation is performed through DNN based object detection model. In last few years, there has been significant advancement in object detection using deep learning techniques. These techniques are fast and robust. We adopted YOLOv3 \cite{redmon2018yolov3} model for hand region segmentation. The model is retrained on the SCUT-Ego-Gesture dataset \cite{wu2017yolse} for hand detection. In the current work, only a single hand region is being detected. The performance of YOLOv3 \cite{redmon2018yolov3} is compared with ground-truth bounding box data given in the SCUT-Ego-Gesture dataset. The performance is compared using the Jaccard similarity index (also known as Intersection over union (IoU)). It is defined as 

\begin{equation}
IoU  = \frac{S_p \cap S_g}{S_p \cup S_g}
\end{equation}
where $S_p$ is the area under the predicted bounding box and $S_g$ is the area under the ground truth bounding box. The hand region segmentation is also performed with the DNN model proposed by Y. Huang et al. \cite{huang2015deepfinger}. We have also designed two ResNet50 \cite{kaiming2016deepresidual} based architectures to perform hand region segmentation. In the first model, hand region segmentation is performed with a direct regression technique, while in the second model, bounding box coordinates are obtained relative to a bounding box (called prior) of the fixed aspect ratio. The performance of each model used for hand region segmentation is furnished in Table \ref{tab:hand-detect}. The YOLOv3 model has an IoU of $0.9046$ on the test set of the SCUT-Ego-Gesture dataset.

\textcolor{black}{The main advantage of YOLOv3 \cite{redmon2018yolov3} over other hand detector used in \cite{huang2015deepfinger, huang2016pointing, wu2017yolse, mukherjee2019fingertip} is its robustness and speed in the detection of hand region. The only requirement of the fingertip detection module in the proposed work is that the cropped image should contain all the visible extend fingers present in the input image. And that is the only dependence of the the proposed method for fingertip detection process on hand detection module. The YOLOv3 \cite{wu2017yolse} is robust enough in locating the hand region from the input image as validated with our experiments (Table \ref{tab:hand-detect}). Therefore, the effect of hand detection process on fingertip detection process is negligible. Some instances of hand region detection using YOLOv3 architecture are presented in Fig. \ref{fig:hand_detect}.}

\subsection{Fingertips detection}
The next step after the hand region segmentation process is the estimation of the fingertips position. $Precision$, $recall$, and $f_1score$ have been used to measure the performance of the proposed algorithm. These metrics are defined in terms of true positive (TP), false-positive (FP), and false-negative (FN), and are given as
\begin{equation}
    precison(p) = \frac{TP}{TP + FP}
\end{equation}

\begin{equation}
    recall(r) = \frac{TP}{TP + FN}
\end{equation}

\begin{equation}
    f_1score = \frac{2 * p * r}{p + r}.
\end{equation}
A detected fingertip point is considered as true positive, if it lies in a circle with radius $\delta$ (a threshold value) with center at the ground-truth. If the predicted fingertip point does not match with any of the ground-truth fingertips, it is treated as false positive. In case fingertip is not predicted or the predicted fingertip lies outside the circle, the ground-truth  fingertip is considered as a false negative. In our experiments, we set the value for $\delta$ either 10 or 15 pixels. The performance comparison for three different backbone models used in our experiments for two different thresholds ($\delta$) values is presented in Table \ref{tab:backbones}. When ResNet50 \cite{he2016deep} is used as a backbone to extract features from the RGB image, the model has the least average pixel error of 2.3552 pixels and the best $f_1score$ of $0.9940$ (threshold $\delta$ is equal to 15 pixels). The fastest processing time is achieved with MobileNetv2 backbone \cite{sandler2018mobilenetv2}, but it is at the cost of a minimal decrease in the performance of the model. 

\begin{table}[!b]
\renewcommand{\arraystretch}{1.3}
\centering
\Large
\caption{Comparative results for various performance metrics on SCUT-Ego-Finger \cite{huang2015deepfinger} dataset.}
\label{tab:egofinger}
\resizebox{\linewidth}{!}{%
\begin{tabular}{|l|c|c|c|c|}
\hline
Algorithm & Backbone Model & \begin{tabular}[c]{@{}c@{}}Actual \\ fingertips\end{tabular} & \begin{tabular}[c]{@{}c@{}}Detected \\ fingertips\end{tabular} & \begin{tabular}[c]{@{}c@{}}Avg. Pixel \\ error (px)\end{tabular} \\ \hhline{=====}
YOLSE \cite{wu2017yolse} & - & \multirow{8}{*}{68513} & 66789 & 14.7556 \\ \cline{1-2} \cline{4-5} 
Pointing Gesture \cite{huang2016pointing} & - &  & 68513 & 6.5066 \\ \cline{1-2} \cline{4-5} 
DeepFinger \cite{huang2015deepfinger} & - &  & 68513 & 6.5445 \\ \cline{1-2} \cline{4-5} \hhline{==~==}
\multirow{2}{*}{FPEFI \cite{purnendu2019fingertips}} & MobileNetV2 &  & 68513 & 6.5292 \\ \cline{2-2} \cline{4-5} 
 & ResNet50 &  & 68513 & 5.8138 \\ \cline{1-2} \cline{4-5} \hhline{==~==}
\multirow{3}{*}{\begin{tabular}[c]{@{}c@{}}Proposed \\ (ABFPE)\end{tabular}} & VGG16 &  & 68511 & 5.3934 \\ \cline{2-2} \cline{4-5} 
 & MobileNetV2 &  & 68510 & 5.1846 \\ \cline{2-2} \cline{4-5} 
 & ResNet50 &  & 68490 & \textbf{5.1334} \\ \hline
\end{tabular}%
}
\end{table}

\begin{figure}[!htbp]
    \centering
    \subfloat[SCUT-Ego-Gesture  dataset \cite{wu2017yolse}]{\includegraphics[width=\linewidth]{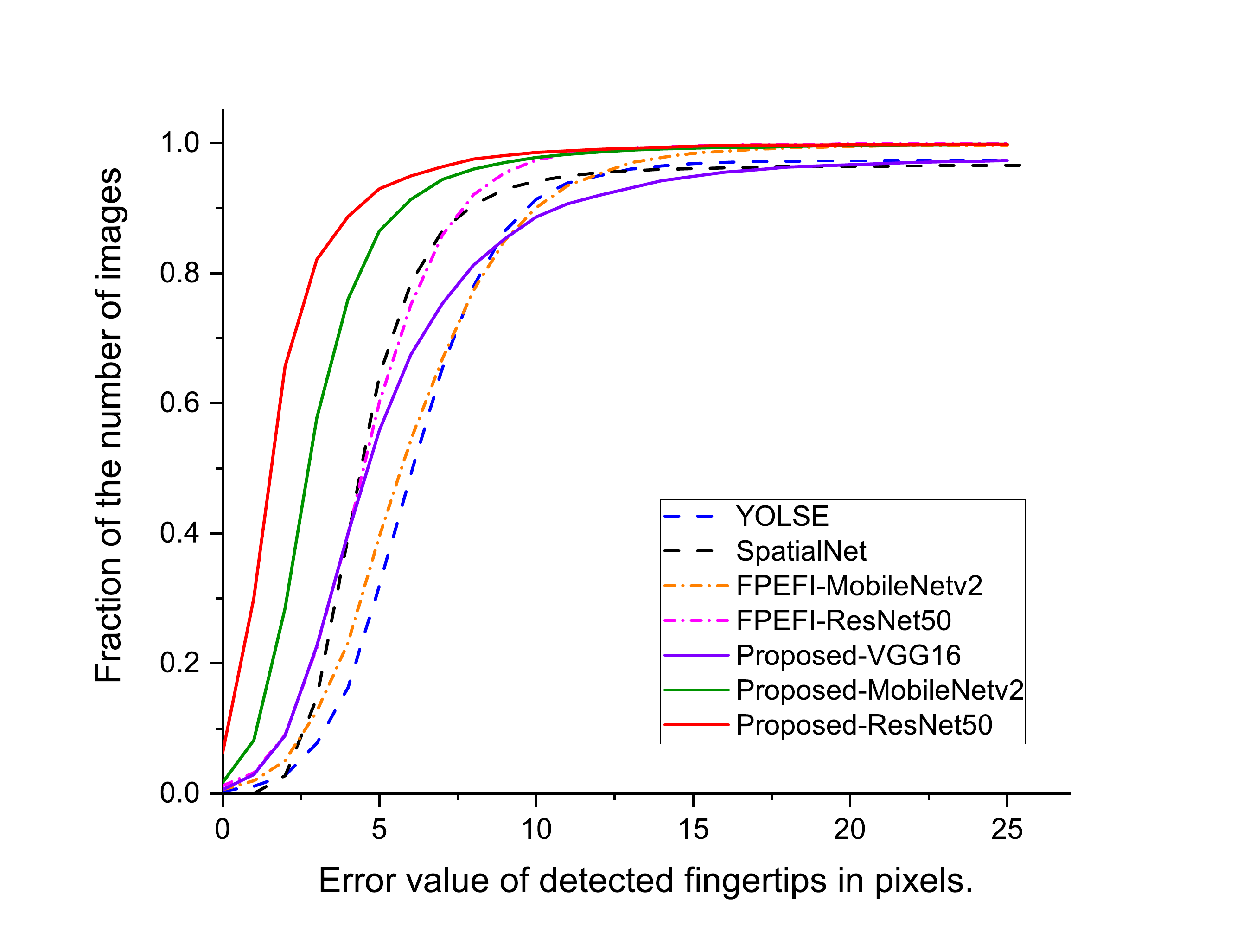}}\\
    \subfloat[SCUT- Ego-Finger  dataset \cite{huang2015deepfinger}]{\includegraphics[width=\linewidth]{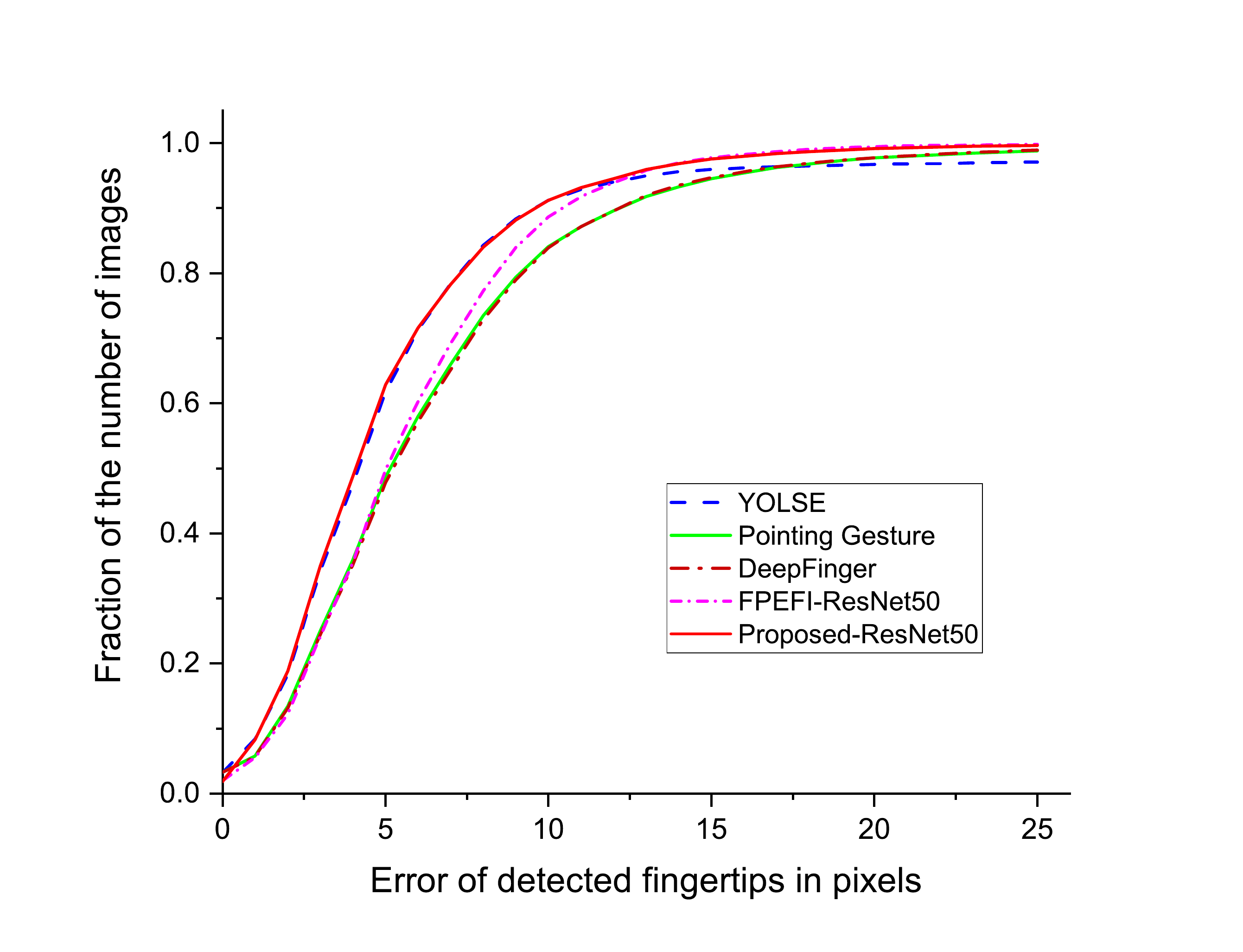}}\\
    \subfloat[HGR dataset \cite{grzejszczak2016hand}]{\includegraphics[width=\linewidth]{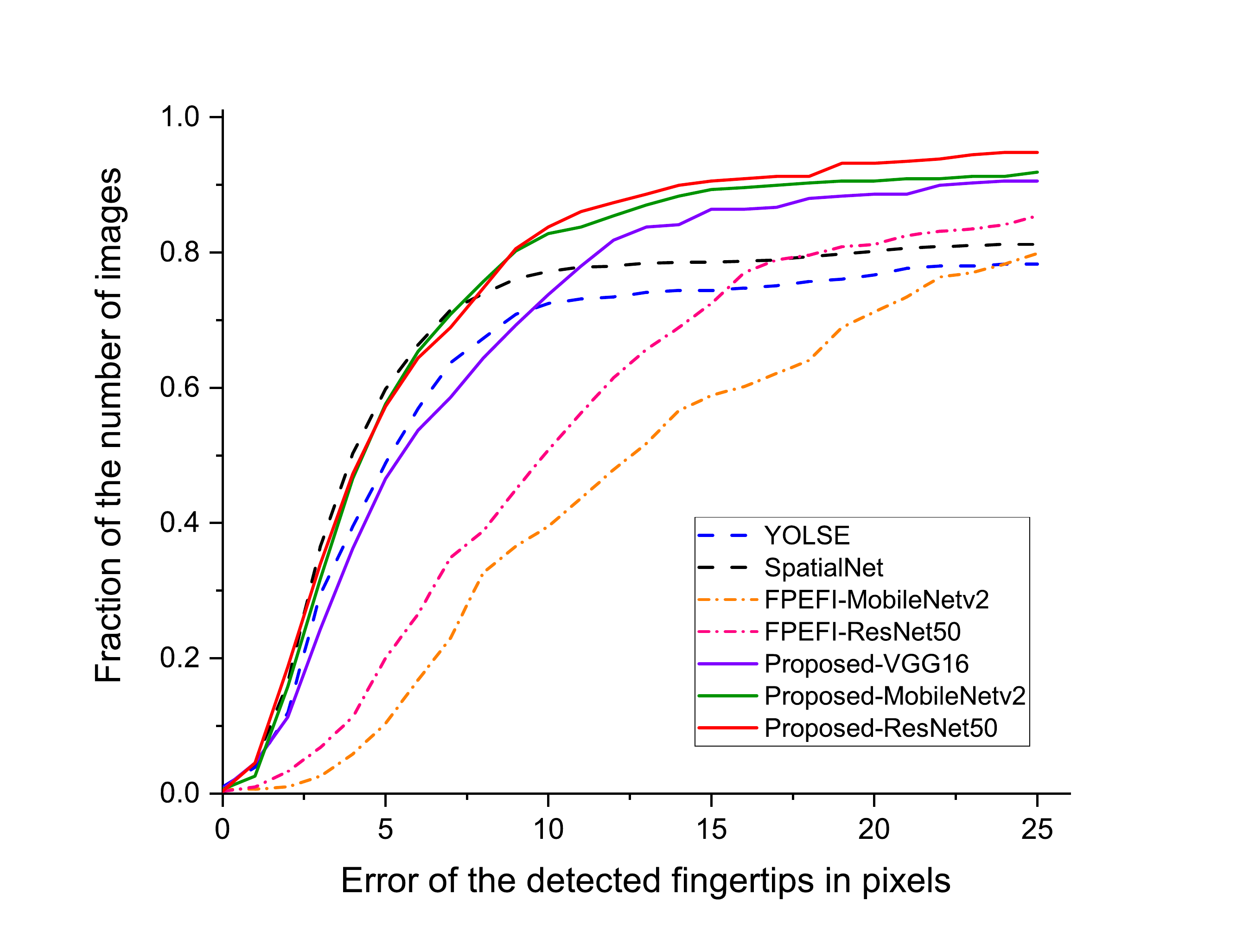}}
    \caption{Comparison of the cumulative distribution curves of the proposed methodology with other existing algorithms on different datasets (a) SCUT-Ego-Gesture dataset \cite{wu2017yolse}, (b) SCUT-Ego-Finger dataset \cite{huang2015deepfinger}, and (c) HGR dataset \cite{grzejszczak2016hand}.}
    \label{fig:cde}
\end{figure}

The performance of the proposed algorithm has been compared with SpatialNet \cite{pfister2015spatialnet}, YOLSE \cite{wu2017yolse}, and FPEFI (Fingertips Position Estimation with Fingers Identification) \cite{purnendu2019fingertips}. The YOLSE and the FPEFI methods present a solution to estimate the position of multiple fingertips on a single RGB image on publicly available datasets. While the SpatialNet \cite{pfister2015spatialnet} was designed to estimate position of seven human  body joints for 2D pose estimation using heatmap regression. The results furnished in Table \ref{tab:egogesture} and Table \ref{tab:hgr} provide comparison of the proposed technique with SpatialNet \cite{pfister2015spatialnet}, YOLSE \cite{wu2017yolse}, and FPEFI \cite{purnendu2019fingertips} on SCUT-Ego-Gesture dataset \cite{wu2017yolse} and HGR datasets \cite{grzejszczak2016hand}, respectively. These experiments are performed with the threshold value ($\delta$) equal to 15. In these datasets, the proposed algorithm (ABFPE) has the least average pixel error as compared to SpatialNet, YOLSE, and FPEFI. The improvement in the estimation of fingertips position is also evident from the qualitative results presented in Fig. \ref{fig:egogesture} and Fig. \ref{fig:hgr} on SCUT-Ego-Gesture dataset \cite{wu2017yolse} and HGR dataset \cite{grzejszczak2016hand}, respectively. The predicted fingertips location is nearest to the ground-truth location in the proposed method. The prediction of SpatialNet and YOLSE has more false negatives or multiple predictions for a ground-truth fingertip location, as shown in first and second row of Fig. \ref{fig:hgr}, respectively. 

The performance of the proposed method is also tested on SCUT-Ego-Finger \cite{huang2015deepfinger} dataset. This dataset comprises of hand images in the egocentric view with only index finger in the extended position. The performance comparison results are furnished in Table \ref{tab:egofinger}. The proposed model has a success rate of $99.96\%$, whereas the YOLSE model \cite{wu2017yolse} can detect $97.48\%$ of fingertips on the total number of the test images.

The performance of the proposed methodology has also been compared in terms of cumulative distribution error (CDE). The CDE calculates the percentages of images that lie under thresholds. The cumulative distribution curve for each of the three datasets has been shown in Fig. \ref{fig:cde}. About $92.98\%$ of the test images from the SCUT-Ego-Gesture dataset \cite{wu2017yolse} have the average pixel error of five pixels or less, which demonstrates the robustness of the proposed method.

\begin{figure}[!t]
    \centering
    \includegraphics[width=\linewidth]{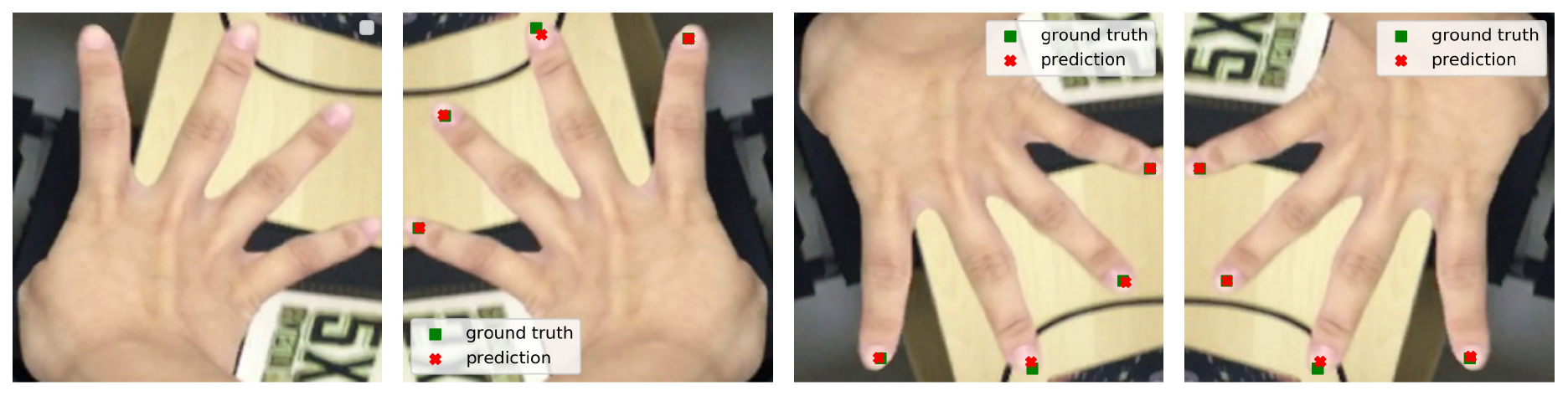}
    \caption{There is no effect on performance of the proposed model due to the rotation of the input image.}
    \label{fig:rotation}
\end{figure}

We also tested the effect of image rotation on the performance of the proposed model and observed that the performance of the proposed algorithm remains unchanged, as shown in Fig. \ref{fig:rotation}. The use of anchor points combined with the capability of DNN is useful in estimating the fingertips position of multiple fingers from a single RGB image. Due to the small network size, the proposed model can perform fingertips estimation in real-time on the GPU enabled system. 

\section{Conclusion}
\label{sec:conclusion}
The algorithm proposed in this paper is focused on the estimation of fingertips position from the single RGB image. With only a small dependence on the hand detection process, the proposed methodology is robust in the estimation of fingertips position of all the visible fingers used in posing a gesture. In comparison to the heatmap-based process, the proposed method has performed well in fingertips detection. The image background factor, illumination variations, skin color, have a minimum effect on the robustness of the proposed technique. Further, the proposed method is rotation invariant. In proposed work, only fingertip(s) location of the extended fingers used while posing a gesture are estimated. In the future, the proposed work will be extended to the estimate fingertip positions of all the visible fingers in an image (extended or non-extended) as well as the detection of other hand-landmark joints to define the hand pose.

\bibliographystyle{IEEEtran}
\bibliography{ref}

\begin{thebibliography}{10}
\providecommand{\url}[1]{#1}
\csname url@samestyle\endcsname
\providecommand{\newblock}{\relax}
\providecommand{\bibinfo}[2]{#2}
\providecommand{\BIBentrySTDinterwordspacing}{\spaceskip=0pt\relax}
\providecommand{\BIBentryALTinterwordstretchfactor}{4}
\providecommand{\BIBentryALTinterwordspacing}{\spaceskip=\fontdimen2\font plus
\BIBentryALTinterwordstretchfactor\fontdimen3\font minus
  \fontdimen4\font\relax}
\providecommand{\BIBforeignlanguage}[2]{{%
\expandafter\ifx\csname l@#1\endcsname\relax
\typeout{** WARNING: IEEEtran.bst: No hyphenation pattern has been}%
\typeout{** loaded for the language `#1'. Using the pattern for}%
\typeout{** the default language instead.}%
\else
\language=\csname l@#1\endcsname
\fi
#2}}
\providecommand{\BIBdecl}{\relax}
\BIBdecl

\bibitem{huang2015deepfinger}
Y.~{Huang}, X.~{Liu}, L.~{Jin}, and X.~{Zhang}, ``{DeepFinger: A Cascade
  Convolutional Neuron Network Approach to Finger Key Point Detection in
  Egocentric Vision with Mobile Camera},'' in \emph{2015 IEEE International
  Conference on Systems, Man, and Cybernetics}, Oct 2015, pp. 2944--2949.

\bibitem{huang2016pointing}
Y.~Huang, X.~Liu, X.~Zhang, and L.~Jin, ``{A Pointing Gesture Based Egocentric
  Interaction System: Dataset, Approach and Application},'' in \emph{2016 IEEE
  Conference on Computer Vision and Pattern Recognition Workshops (CVPRW)},
  June 2016, pp. 370--377.

\bibitem{mukherjee2019fingertip}
S.~Mukherjee, S.~A. Ahmed, D.~P. Dogra, S.~Kar, and P.~P. Roy, ``{Fingertip
  detection and tracking for recognition of air-writing in videos},''
  \emph{Expert Systems with Applications}, vol. 136, pp. 217--229, 2019.

\bibitem{buchmann2004fingartips}
V.~Buchmann, S.~Violich, M.~Billinghurst, and A.~Cockburn, ``{FingARtips:
  Gesture based direct manipulation in Augmented Reality},'' in
  \emph{Proceedings of the 2nd international conference on Computer graphics
  and interactive techniques in Australasia and South East Asia}.\hskip 1em
  plus 0.5em minus 0.4em\relax ACM, 2004, pp. 212--221.

\bibitem{liu2019grasp}
C.~{Liu}, B.~{Fang}, F.~{Sun}, X.~{Li}, and W.~{Huang}, ``{Learning to Grasp
  Familiar Objects Based on Experience and Objects’ Shape Affordance},''
  \emph{IEEE Transactions on Systems, Man, and Cybernetics: Systems}, vol.~49,
  no.~12, pp. 2710--2723, Dec 2019.

\bibitem{mistry2009wearable}
\BIBentryALTinterwordspacing
P.~Mistry and P.~Maes, ``{SixthSense: A Wearable Gestural Interface},'' in
  \emph{ACM SIGGRAPH ASIA 2009 Sketches}, ser. SIGGRAPH ASIA '09.\hskip 1em
  plus 0.5em minus 0.4em\relax New York, NY, USA: ACM, 2009, pp. 11:1--11:1.
  [Online]. Available: \url{http://doi.acm.org/10.1145/1667146.1667160}
\BIBentrySTDinterwordspacing

\bibitem{billinghurst2014hands}
M.~{Billinghurst}, T.~{Piumsomboon}, and H.~{Bai}, ``{Hands in Space: Gesture
  Interaction with Augmented-Reality Interfaces},'' \emph{IEEE Computer
  Graphics and Applications}, vol.~34, no.~1, pp. 77--80, Jan 2014.

\bibitem{zhang2013new}
X.~Zhang, Z.~Ye, L.~Jin, Z.~Feng, and S.~Xu, ``{A new writing experience:
  Finger writing in the air using a Kinect sensor},'' \emph{IEEE MultiMedia},
  vol.~20, no.~4, pp. 85--93, 2013.

\bibitem{choi2018co}
O.~{Choi}, Y.~{Son}, H.~{Lim}, and S.~C. {Ahn}, ``{Co-Recognition of Multiple
  Fingertips for Tabletop Human-Projector Interaction},'' \emph{IEEE Trans.
  Multimedia}, vol.~21, no.~6, pp. 1487--1498, June 2019.

\bibitem{modanwal2018robustwrist}
G.~Modanwal and K.~Sarawadekar, ``A robust wrist point detection algorithm
  using geometric features,'' \emph{Pattern Recognition Letters}, vol. 110, pp.
  72--78, 2018.

\bibitem{modanwal2018robusthand}
G.~Modanwal, P.~Mishra, and K.~Sarawadekar, ``{A Robust Algorithm for
  Hand-Forearm Segmentation},'' in \emph{Proceedings of the 2018 International
  Conference on Image and Graphics Processing}.\hskip 1em plus 0.5em minus
  0.4em\relax ACM, 2018, pp. 102--105.

\bibitem{dorfmuller2001finger}
K.~{Dorfmuller-Ulhaas} and D.~{Schmalstieg}, ``{Finger tracking for interaction
  in augmented environments},'' in \emph{Proceedings IEEE and ACM International
  Symposium on Augmented Reality}, Oct 2001, pp. 55--64.

\bibitem{zhang2005interaction}
R.~Zhang and X.~Zhang, ``Interaction method based on data glove in virtual
  environment,'' \emph{Computer Engineering}, vol.~12, 2005.

\bibitem{nakamura2008double}
T.~Nakamura, S.~Takahashi, and J.~Tanaka, ``{Double-Crossing: A New Interaction
  Technique for Hand Gesture Interfaces},'' in \emph{Asia-Pacific Conference on
  Computer Human Interaction}.\hskip 1em plus 0.5em minus 0.4em\relax Springer,
  2008, pp. 292--300.

\bibitem{lee2012robust}
{Lae-Kyoung Lee}, {Su-Yong An}, and {Se-Young Oh}, ``{Robust fingertip
  extraction with improved skin color segmentation for finger gesture
  recognition in Human-robot interaction},'' in \emph{2012 IEEE Congress on
  Evolutionary Computation}, June 2012, pp. 1--7.

\bibitem{wu2016robust}
G.~{Wu} and W.~{Kang}, ``{Robust Fingertip Detection in a Complex
  Environment},'' \emph{IEEE Trans. Multimedia}, vol.~18, no.~6, pp. 978--987,
  June 2016.

\bibitem{liang2014parsing}
H.~{Liang}, J.~{Yuan}, and D.~{Thalmann}, ``{Parsing the Hand in Depth
  Images},'' \emph{IEEE Trans. Multimedia}, vol.~16, no.~5, pp. 1241--1253, Aug
  2014.

\bibitem{ren2013robust}
Z.~{Ren}, J.~{Yuan}, J.~{Meng}, and Z.~{Zhang}, ``{Robust Part-Based Hand
  Gesture Recognition Using Kinect Sensor},'' \emph{IEEE Trans. Multimedia},
  vol.~15, no.~5, pp. 1110--1120, Aug 2013.

\bibitem{wang2015superpixel}
C.~{Wang}, Z.~{Liu}, and S.~{Chan}, ``{Superpixel-Based Hand Gesture
  Recognition With Kinect Depth Camera},'' \emph{IEEE Trans. on Multimedia},
  vol.~17, no.~1, pp. 29--39, Jan 2015.

\bibitem{wu2017yolse}
W.~{Wu}, C.~{Li}, Z.~{Cheng}, X.~{Zhang}, and L.~{Jin}, ``{YOLSE: Egocentric
  Fingertip Detection from Single RGB Images},'' in \emph{2017 IEEE
  International Conference on Computer Vision Workshops (ICCVW)}, Oct 2017, pp.
  623--630.

\bibitem{wang2020srhandnet}
Y.~{Wang}, B.~{Zhang}, and C.~{Peng}, ``{SRHandNet: Real-Time 2D Hand Pose
  Estimation With Simultaneous Region Localization},'' \emph{IEEE Transactions
  on Image Processing}, vol.~29, pp. 2977--2986, 2020.

\bibitem{ren2015faster}
S.~Ren, K.~He, R.~Girshick, and J.~Sun, ``{Faster R-CNN: Towards Real-Time
  Object Detection with Region Proposal Networks},'' in \emph{Advances in
  neural information processing systems}, 2015, pp. 91--99.

\bibitem{redmon2018yolov3}
J.~Redmon and A.~Farhadi, ``{YOLO}v3: An incremental improvement,'' \emph{arXiv
  preprint arXiv:1804.02767}, 2018.

\bibitem{baldauf2011markerless}
M.~Baldauf, S.~Zambanini, P.~Fr\"{o}hlich, and P.~Reichl, ``{Markerless Visual
  Fingertip Detection for Natural Mobile Device Interaction},'' in
  \emph{Proceedings of the 13th International Conference on Human Computer
  Interaction with Mobile Devices and Services}, ser. MobileHCI ’11.\hskip
  1em plus 0.5em minus 0.4em\relax New York, NY, USA: Association for Computing
  Machinery, 2011, p. 539–544.

\bibitem{lee2011vision}
D.~Lee and S.~Lee, ``{Vision-Based Finger Action Recognition by Angle Detection
  and Contour Analysis},'' \emph{ETRI journal}, vol.~33, no.~3, pp. 415--422,
  2011.

\bibitem{kulshreshth2013poster}
A.~Kulshreshth, C.~Zorn, and J.~J. LaViola, ``{Poster: Real-time markerless
  kinect based finger tracking and hand gesture recognition for HCI},'' in
  \emph{2013 IEEE Symposium on 3D User Interfaces (3DUI)}.\hskip 1em plus 0.5em
  minus 0.4em\relax IEEE, 2013, pp. 187--188.

\bibitem{purnendu2019fingertips}
P.~{Mishra} and K.~{Sarawadekar}, ``{Fingertips Detection in Egocentric Video
  Frames using Deep Neural Networks},'' in \emph{2019 International Conference
  on Image and Vision Computing New Zealand (IVCNZ)}, Dec 2019, pp. 1--6.

\bibitem{redmon2017yolo9000}
J.~Redmon and A.~Farhadi, ``Yolo9000: {B}etter, {F}aster, {S}tronger,'' in
  \emph{Proceedings of the IEEE conference on computer vision and pattern
  recognition}, 2017, pp. 7263--7271.

\bibitem{liu2016ssd}
W.~Liu, D.~Anguelov, D.~Erhan, C.~Szegedy, S.~Reed, C.-Y. Fu, and A.~C. Berg,
  ``{SSD}: Single shot multibox detector,'' in \emph{European conference on
  computer vision}.\hskip 1em plus 0.5em minus 0.4em\relax Springer, 2016, pp.
  21--37.

\bibitem{he2016deep}
K.~He, X.~Zhang, S.~Ren, and J.~Sun, ``Deep residual learning for image
  recognition,'' in \emph{Proceedings of the IEEE conference on computer vision
  and pattern recognition}, 2016, pp. 770--778.

\bibitem{vgg16}
K.~Simonyan and A.~Zisserman, ``Very deep convolutional networks for
  large-scale image recognition,'' \emph{CoRR}, vol. abs/1409.1556, 2014.

\bibitem{sandler2018mobilenetv2}
M.~Sandler, A.~Howard, M.~Zhu, A.~Zhmoginov, and L.-C. Chen, ``{Mobilenetv2:
  Inverted residuals and linear bottlenecks},'' in \emph{Proceedings of the
  IEEE Conference on Computer Vision and Pattern Recognition}, 2018, pp.
  4510--4520.

\bibitem{krizhevsky2012imagenet}
A.~Krizhevsky, I.~Sutskever, and G.~E. Hinton, ``Imagenet classification with
  deep convolutional neural networks,'' in \emph{Advances in neural information
  processing systems}, 2012, pp. 1097--1105.

\bibitem{liu2015fingertip}
X.~Liu, Y.~Huang, X.~Zhang, and L.~Jin, ``Fingertip in the eye: A cascaded cnn
  pipeline for the real-time fingertip detection in egocentric videos,''
  \emph{arXiv preprint arXiv:1511.02282}, 2015.

\bibitem{kawulok2014self}
M.~Kawulok, J.~Kawulok, J.~Nalepa, and B.~Smolka, ``{Self-adaptive algorithm
  for segmenting skin regions},'' \emph{EURASIP Journal on Advances in Signal
  Processing}, vol. 2014, no.~1, p. 170, 2014.

\bibitem{nalepa2014fast}
J.~Nalepa and M.~Kawulok, ``{Fast and accurate hand shape classification},'' in
  \emph{International conference: beyond databases, architectures and
  structures}.\hskip 1em plus 0.5em minus 0.4em\relax Springer, 2014, pp.
  364--373.

\bibitem{grzejszczak2016hand}
{Grzejszczak, Tomasz and Kawulok, Michal and Galuszka, Adam}, ``{Hand landmarks
  detection and localization in color images},'' \emph{Multimedia Tools and
  Applications}, vol.~75, no.~23, pp. 16\,363--16\,387, 2016.

\bibitem{huber1992robust}
P.~J. Huber, ``{Robust Estimation of a Location Parameter},'' in
  \emph{Breakthroughs in statistics}.\hskip 1em plus 0.5em minus 0.4em\relax
  Springer, 1992, pp. 492--518.

\bibitem{purnendu2019polynomial}
P.~{Mishra} and K.~{Sarawadekar}, ``{Polynomial Learning Rate Policy with Warm
  Restart for Deep Neural Network},'' in \emph{TENCON 2019 - 2019 IEEE Region
  10 Conference (TENCON)}, Oct 2019, pp. 2087--2092.

\bibitem{chollet2015keras}
F.~Chollet \emph{et~al.}, ``Keras,'' \url{https://keras.io}, 2015.

\bibitem{pfister2015spatialnet}
T.~{Pfister}, J.~{Charles}, and A.~{Zisserman}, ``{Flowing ConvNets for Human
  Pose Estimation in Videos},'' in \emph{2015 IEEE International Conference on
  Computer Vision (ICCV)}, Dec 2015, pp. 1913--1921.

\bibitem{kaiming2016deepresidual}
K.~He, X.~Zhang, S.~Ren, and J.~Sun, ``Identity mappings in deep residual
  networks,'' in \emph{Computer Vision -- ECCV 2016}, B.~Leibe, J.~Matas,
  N.~Sebe, and M.~Welling, Eds.\hskip 1em plus 0.5em minus 0.4em\relax Cham:
  Springer International Publishing, 2016, pp. 630--645.

\end{thebibliography}

\end{document}